\documentclass[11pt,a4paper]{article}
\usepackage[hyperref]{acl2020}
\usepackage{times}
\usepackage{latexsym}

\usepackage{microtype}
\usepackage{graphicx}
\usepackage{multirow}
\usepackage{graphicx}
\usepackage{booktabs}
\usepackage{amsthm}

\newtheorem{myDef}{Definition}
\newtheorem{myHypo}{Hypothesis}

\aclfinalcopy % Uncomment this line for the final submission
\def\aclpaperid{378} %  Enter the acl Paper ID here

\title{FastBERT: a Self-distilling BERT with Adaptive Inference Time}

\author{\\
Weijie Liu\textsuperscript{\rm 1,2}, 
Peng Zhou\textsuperscript{\rm 2}, 
Zhe Zhao\textsuperscript{\rm 2}, 
Zhiruo Wang\textsuperscript{\rm 3},
Haotang Deng\textsuperscript{\rm 2}
and Qi Ju\textsuperscript{\rm 2,\thanks{$^*$Corresponding author: Qi Ju (damonju@tencent.com)}}\\ 
\textsuperscript{\rm 1}Peking University, Beijing, China \\
\textsuperscript{\rm 2}Tencent Research, Beijing, China\\
\textsuperscript{\rm 3}Beijing Normal University, Beijing, China \\
\small{dataliu@pku.edu.cn, \{rickzhou, nlpzhezhao, haotangdeng, damonju\}@tencent.com}, SherronWang@gmail.com\\
}

\date{}

\begin{document}
\maketitle
\begin{abstract}

Pre-trained language models like BERT have proven to be highly performant. However, they are often computationally expensive in many practical scenarios, for such heavy models can hardly be readily implemented with limited resources. To improve their efficiency with an assured model performance, we propose a novel speed-tunable FastBERT with adaptive inference time. The speed at inference can be flexibly adjusted under varying demands, while redundant calculation of samples is avoided. Moreover, this model adopts a unique self-distillation mechanism at fine-tuning, further enabling a greater computational efficacy with minimal loss in performance. Our model achieves promising results in twelve English and Chinese datasets. It is able to speed up by a wide range from 1 to 12 times than BERT if given different speedup thresholds to make a speed-performance tradeoff.

\end{abstract}

\section{Introduction}

Last two years have witnessed significant improvements brought by language pre-training, such as BERT \citep{devlin2019bert}, GPT \citep{radford2018improving}, and XLNet \citep{yang2019xlnet}. By pre-training on unlabeled corpus and fine-tuning on labeled ones, BERT-like models achieved huge gains on many Natural Language Processing tasks.

Despite this gain in accuracy, these models have greater costs in computation and slower speed at inference, which severely impairs their practicalities. Actual settings, especially with limited time and resources in the industry, can hardly enable such models into operation. For example, in tasks like sentence matching and text classification, one often requires to process billions of requests per second. What's more, the number of requests varies with time. In the case of an online shopping site, the number of requests during the holidays is five to ten times more than that of the workdays. A large number of servers need to be deployed to enable BERT in industrial settings, and many spare servers need to be reserved to cope with the peak period of requests, demanding huge costs.

To improve their usability, many attempts in model acceleration have been made, such as quantinization \citep{gong2014compressing}, weights pruning \citep{han2015learning}, and knowledge distillation (KD) \citep{romero2014fitnets}. As one of the most popular methods, KD requires additional smaller student models that depend entirely on the bigger teacher model and trade task accuracy for ease in computation \citep{hinton2015distilling}. 
Reducing model sizes to achieve acceptable speed-accuracy balances, however, can only solve the problem halfway, for the model is still set as fixated, rendering them unable to cope with drastic changes in request amount.

By inspecting many NLP datasets \citep{wang2018glue}, we discerned that the samples have different levels of difficulty. Heavy models may over-calculate the simple inputs, while lighter ones are prone to fail in complex samples. As recent studies \citep{kovaleva2019revealing} have shown redundancy in pre-training models, it is useful to design a one-size-fits-all model that caters to samples with varying complexity and gains computational efficacy with the least loss of accuracy. 

Based on this appeal, we propose FastBERT, a pre-trained model with a sample-wise adaptive mechanism. It can adjust the number of executed layers dynamically to reduce computational steps. This model also has a unique self-distillation process that requires minimal changes to the structure, achieving faster yet as accurate outcomes within a single framework. Our model not only reaches a comparable speedup (by 2 to 11 times) to the BERT model, but also attains competitive accuracy in comparison to heavier pre-training models.

Experimental results on six Chinese and six English NLP tasks have demonstrated that FastBERT achieves a huge retrench in computation with very little loss in accuracy. The main contributions of this paper can be summarized as follows:

\begin{itemize}
	\item This paper proposes a practical speed-tunable BERT model, namely FastBERT, that balances the speed and accuracy in the response of varying request amounts;
	\item The sample-wise adaptive mechanism and the self-distillation mechanism are combined to improve the inference time of NLP model for the first time. Their efficacy is verified on twelve NLP datasets;
	\item The code is publicly available at \url{https://github.com/autoliuweijie/FastBERT}.
\end{itemize}

\section{Related work}

BERT \citep{devlin2019bert} can learn universal knowledge from mass unlabeled data and produce more performant outcomes. Many works have followed: RoBERTa \citep{liu2019roberta} that uses larger corpus and longer training steps. T5 \citep{raffel2019exploring} that scales up the model size even more. UER \citep{zhao2019uer} pre-trains BERT in different Chinese corpora. K-BERT \citep{liu2019k} injects knowledge graph into BERT model. These models achieve increased accuracy with heavier settings and even more data.

However, such unwieldy sizes are often hampered under stringent conditions. To be more specific, BERT-base contains 110 million parameters by stacking twelve Transformer blocks \citep{vaswani2017attention}, while BERT-large expands its size to even 24 layers. ALBERT \citep{lan2019albert} shares the parameters of each layer to reduce the model size. Obviously, the inference speed for these models would be much slower than classic architectures (e.g., CNN \citep{kim-2014-convolutional}, RNN \citep{wang-2018-disconnected}, etc). We think a large proportion of computation is caused by redundant calculation.

\begin{figure}[t]
\centering
\includegraphics[width=0.7\columnwidth]{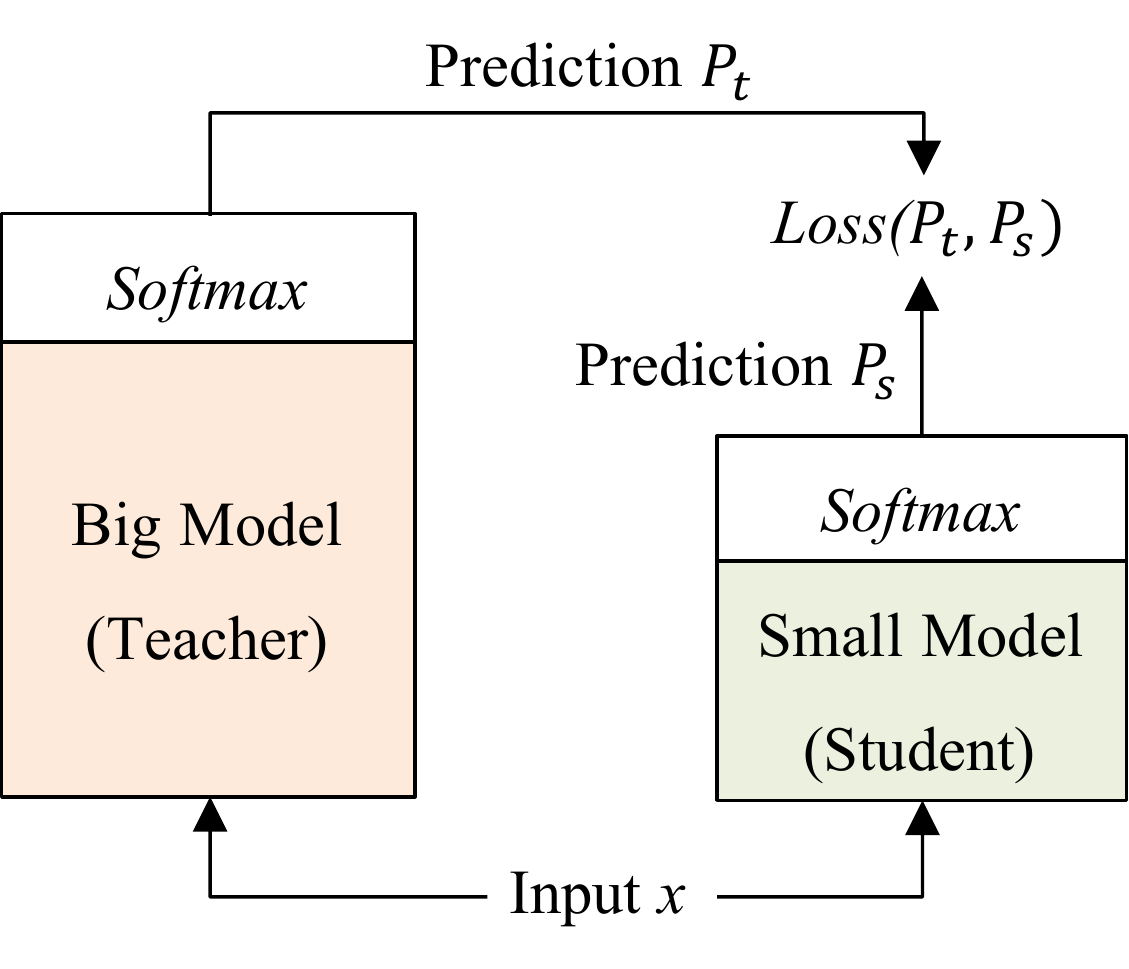}
\caption{Classic knowledge distillation approach: Distill a small model using a separate big model.}
\label{fig:classic_distillation}
\end{figure}

\textbf{Knowledge distillation}: Many attempts have been made to distill heavy models (teachers) into their lighter counterparts (students). PKD-BERT \citep{sun2019patient} adopts an incremental extraction process that learns generalizations from intermediate layers of the teacher model. TinyBERT \citep{jiao2019tinybert} performs a two-stage learning involving both general-domain pre-training and task-specific fine-tuning. DistilBERT \citep{sanh2019distilbert} further leveraged the inductive bias within large models by introducing a triple loss. As shown in Figure \ref{fig:classic_distillation}, student model often require a separated structure, whose effect however, depends mainly on the gains of the teacher. They are as indiscriminate to individual cases as their teachers, and only get faster in the cost of degraded performance. 

\textbf{Adaptive inference}: Conventional approaches in adaptive computations are performed token-wise or patch-wise, who either adds recurrent steps to individual tokens \citep{graves2016adaptive} or dynamically adjusts the number of executed layers inside discrete regions of images \citep{teerapittayanon2016branchynet, figurnov2017spatially}. To the best of our knowledge, there has been no work in applying adaptive mechanisms to NLP pre-training language models for efficiency improvements so far.

\begin{figure*}[t]
\centering
\includegraphics[width=.95\textwidth]{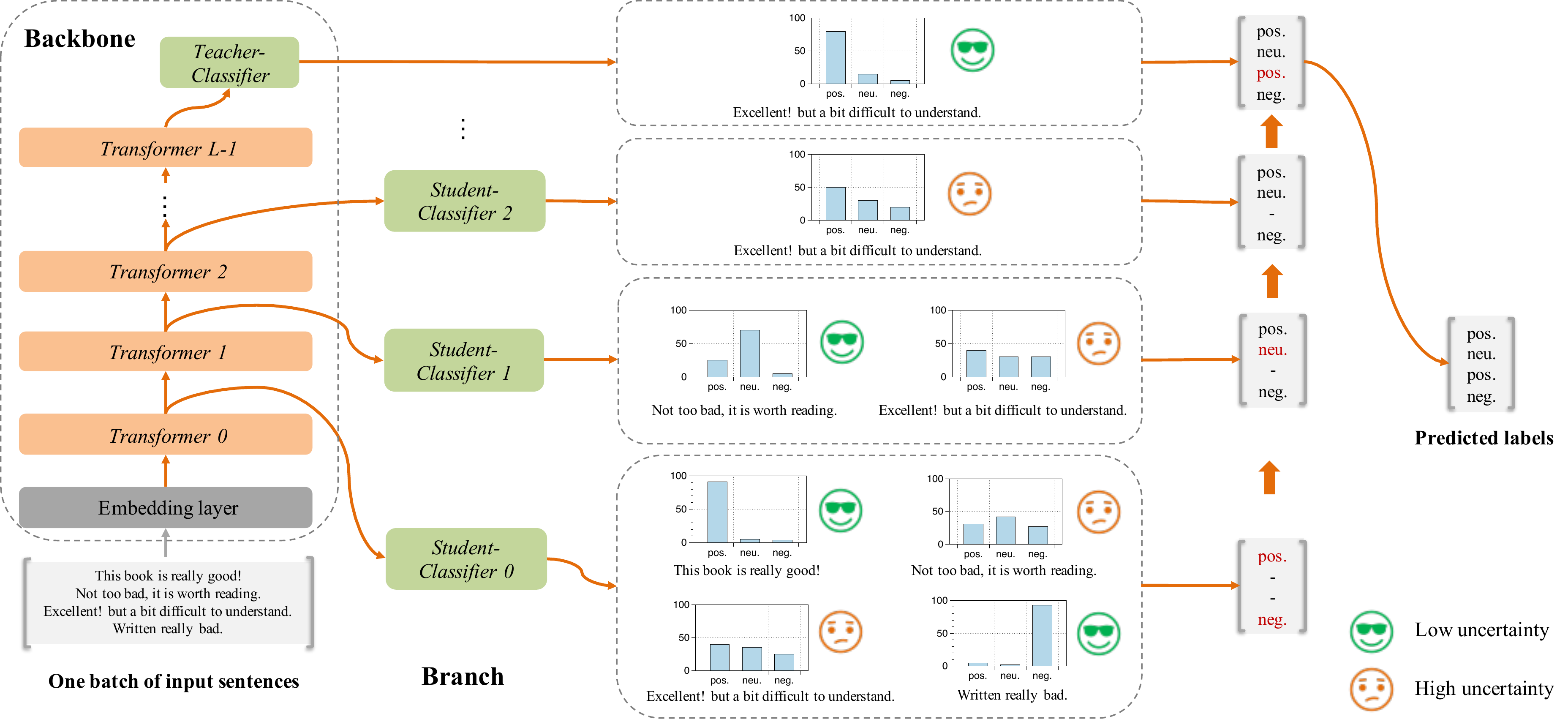}
\caption{The inference process of FastBERT, where the number of executed layers with each sample varies based on its complexity. This illustrates a sample-wise adaptive mechanism. Taking a batch of inputs ($batch\_size=4$) as an example, the \textit{Transformer0} and \textit{Student-classifier0} inferred their labels as probability distributions and calculate the individual uncertainty. Cases with low uncertainty are immediately removed from the batch, while those with higher uncertainty are sent to the next layer for further inference.}
\label{fig:model_atchitecture}
\end{figure*}

\section{Methodology}
\label{sec:methodology}

Distinct to the above efforts, our approach fusions the adaptation and distillation into a novel speed-up approach, shown in Figure \ref{fig:model_atchitecture}, achieving competitive results in both accuracy and efficiency.

\subsection{Model architecture}
\label{sec:model_architecture}

As shown in Figure \ref{fig:model_atchitecture}, FastBERT consists of backbone and branches. The backbone is built upon 12-layers Transformer encoder with an additional teacher-classifier, while the branches include student-classifiers which are appended to each Transformer output to enable early outputs.

\subsubsection{Backbone}
\label{sec:Backbone}

The backbone consists of three parts: the embedding layer, the encoder containing stacks of Transformer blocks \citep{vaswani2017attention}, and the teacher classifier. The structure of the embedding layer and the encoder conform with those of BERT \citep{devlin2019bert}. 
Given the sentence length $n$, an input sentence $s=[w_0, w_1, ... w_n]$ will be transformed by the embedding layers to a sequence of vector representations $e$ like (\ref{eq:embedding}),
\begin{equation}
\label{eq:embedding}
e = Embedding(s),
\end{equation}
where $e$ is the summation of word, position, and segment embeddings. Next, the transformer blocks in the encoder performs a layer-by-layer feature extraction as (\ref{eq:encoder}),
\begin{equation}
\label{eq:encoder}
h_i = Transformer\_i(h_{i-1}),
\end{equation}
where $h_i$ ($i=-1,0,1,...,L-1$) is the output features at the $i$th layer, and $h_{-1}=e$. $L$ is the number of Transformer layers.

Following the final encoding output is a teacher classifier that extracts in-domain features for downstream inferences. It has a fully-connected layer narrowing the dimension from $768$ to $128$, a self-attention joining a fully-connected layer without changes in vector size, and a fully-connected layer with a $softmax$ function projecting vectors to an $N$-class indicator $p_t$ as in (\ref{eq:teacher}), where $N$ is the task-specific number of classes.
\begin{equation}
\label{eq:teacher}
p_t = Teacher\_Classifier(h_{L-1}).
\end{equation} 

\subsubsection{Branches}
\label{sec:branch}

To provide FastBERT with more adaptability, multiple branches, i.e. the student classifiers, in the same architecture with the teacher are added to the output of each Transformer block to enable early outputs, especially in those simple cases. The student classifiers can be described as (\ref{eq:student}),
\begin{equation}
\label{eq:student}
p_{s_i} = Student\_Classifier\_i(h_{i}).
\end{equation}

The student classifier is designed carefully to balance model accuracy and inference speed, for simple networks may impair the performance, while a heavy attention module severely slows down the inference speed. Our classifier has proven to be lighter with ensured competitive accuracy, detailed verifications are showcased in Section \ref{sec:flops_analysis}.

\subsection{Model training}

FastBERT requires respective training steps for the backbone and the student classifiers. The parameters in one module is always frozen while the other module is being trained. The model is trained in preparation for downstream inference with three steps: the major backbone pre-training, entire backbone fine-tuning, and self-distillation for student classifiers. 

\subsubsection{Pre-training}
\label{sec:Pre-training}

The pre-training of backbone resembles that of BERT in the same way that our backbone resembles BERT. Any pre-training method used for BERT-like models (e.g., BERT-WWM \citep{cui2019pre}, RoBERTa \citep{liu2019roberta}, and ERNIE \citep{sun2019ernie}) can be directly applied. Note that the teacher classifier, as it is only used for inference, stays unaffected at this time.
Also conveniently, FastBERT does not even need to perform pre-training by itself, for it can load high-quality pre-trained models freely.

\subsubsection{Fine-tuning for backbone}
\label{sec:Fine-tune}

For each downstream task, we plug in the task-specific data into the model, fine-tuning both the major backbone and the teacher classifier. The structure of the teacher classifier is as previously described. At this stage, all student classifiers are not enabled. 

\subsubsection{Self-distillation for branch}
\label{sec:self_distillation}

With the backbone well-trained for knowledge extraction, its output, as a high-quality soft-label containing both the original embedding and the generalized knowledge, is distilled for training student classifiers. 
As student are mutually independent, their predictions $p_s$ are compared with the teacher soft-label $p_t$ respectively, with the differences measured by KL-Divergence in (\ref{eq:kl}),
\begin{equation}
\label{eq:kl}
D_{KL}(p_s, p_t) =\sum_{i=1}^{N}p_s(i)\cdot \log \frac{p_s(i)}{p_t(j)}.
\end{equation}
As there are $L-1$ student classifiers in the FastBERT, the sum of their KL-Divergences is used as the total loss for self-distillation, which is formulated in (\ref{eq:loss}),
\begin{equation}
\label{eq:loss}
Loss(p_{s_0},...,p_{s_{L-2}}, p_t) =\sum_{i=0}^{L-2}D_{KL}(p_{s_{i}}, p_t),
\end{equation}
where $p_{s_i}$ refers to the probability distribution of the output from \textit{student-classifier i}. 

Since this process only requires the teacher‘s output, we are free to use an unlimited number of unlabeled data, instead of being restricted to the labeled ones. This provides us with sufficient resources for self-distillation, which means we can always improve the student performance as long as the teacher allows.
Moreover, our method differs from the previous distillation method, for the teacher and student outputs lie within the same model. This learning process does not require additional pre-training structures, making the distillation entirely a learning process by self.

\begin{table}[b]
\small
\caption{FLOPs of each operation within the FastBERT (M = Million, $N$ = the number of labels).}
\label{tab:operation_flops}
\begin{tabular}{cccc}
\toprule
\textbf{Operation}           & \textbf{Sub-operation}                                               & \textbf{FLOPs} & \textbf{Total FLOPs}     \\  \midrule
\multirow{2}{*}{Transformer} & \begin{tabular}[c]{@{}c@{}}Self-attention\\ ($768\rightarrow768$)\end{tabular}   & 603.0M         & \multirow{2}{*}{1809.9M} \\ 
                             & \begin{tabular}[c]{@{}c@{}}Feedforward\\ ($768\rightarrow3072$\\$\rightarrow768$)\end{tabular} & 1207.9M        &                          \\ \midrule
\multirow{4}{*}{Classifier}  & \begin{tabular}[c]{@{}c@{}}Fully-connect\\ ($768\rightarrow128$)\end{tabular}    & 25.1M          & \multirow{4}{*}{46.1M}   \\
                             & \begin{tabular}[c]{@{}c@{}}Self-attention\\ ($128\rightarrow128$)\end{tabular}   & 16.8M          &                          \\
                             & \begin{tabular}[c]{@{}c@{}}Fully-connect\\ ($128\rightarrow128$)\end{tabular}    & 4.2M           &                          \\
                             & \begin{tabular}[c]{@{}c@{}}Fully-connect\\ ($128\rightarrow N$)\end{tabular}      & -              &                          \\ \bottomrule
\end{tabular}
\end{table}

\subsection{Adaptive inference}
\label{sec:self_adaptive_inference}

With the above steps, FastBERT is well-prepared to perform inference in an adaptive manner, which means we can adjust the number of executed encoding layers within the model according to the sample complexity. 

At each Transformer layer, we measure for each sample on whether the current inference is credible enough to be terminated. 

Given an input sequence, the uncertainty of a student classifier's output $p_s$ is computed with a normalized entropy in (\ref{eq:uncertainty}),
\begin{equation}
\label{eq:uncertainty}
Uncertainty = \frac{\sum_{i=1}^{N}p_s(i) \log p_s(i)}{\log\frac{1}{N}},
\end{equation}
where $p_s$ is the distribution of output probability, and N is the number of labeled classes.

With the definition of the uncertainty, we make an important hypothesis.

\begin{myHypo}
	LUHA: the Lower the Uncertainty, the Higher the Accuracy.
\end{myHypo} 

\begin{myDef}
	Speed: The threshold to distinguish high and low uncertainty.
\end{myDef}

LUHA is verified in Section \ref{hypothesis_verification}. Both \textit{Uncertainty} and \textit{Speed} range between $0$ and $1$. The adaptive inference mechanism can be described as: At each layer of FastBERT, the corresponding \textit{student classifier} will predict the label of each sample with measured \textit{Uncertainty}. Samples with \textit{Uncertainty} below the \textit{Speed} will be sifted to early outputs, while samples with \textit{Uncertainty} above the \textit{Speed} will move on to the next layer.

Intuitively, with a higher \textit{Speed}, fewer samples will be sent to higher layers, and overall inference speed will be faster, and vice versa. Therefore, \textit{Speed} can be used as a halt value for weighing the inference accuracy and efficiency.

\begin{table*}[ht]
\caption{Comparison of accuracy (Acc.) and FLOPs (speedup) between FastBERT and Baselines in six Chinese datasets and six English datasets.}
\label{tab:comparison}
\scriptsize
\begin{tabular}{c|cc|cc|cc|cc|cc|cc}
\toprule
\multirow{2}{*}{\textbf{\begin{tabular}[c]{@{}c@{}}Dataset/\\ Model\end{tabular}}} & \multicolumn{2}{c|}{\textbf{ChnSentiCorp}}                             & \multicolumn{2}{c|}{\textbf{Book review}}                           & \multicolumn{2}{c|}{\textbf{Shopping review}}                           & \multicolumn{2}{c|}{\textbf{LCQMC}}                               & \multicolumn{2}{c|}{\textbf{Weibo}}                               & \multicolumn{2}{c}{\textbf{THUCNews}}                            \\
                                                                                   & Acc.  & \begin{tabular}[c]{@{}c@{}}FLOPs\\ (speedup)\end{tabular} & Acc.  & \begin{tabular}[c]{@{}c@{}}FLOPs\\ (speedup)\end{tabular} & Acc.  & \begin{tabular}[c]{@{}c@{}}FLOPs\\ (speedup)\end{tabular} & Acc.  & \begin{tabular}[c]{@{}c@{}}FLOPs\\ (speedup)\end{tabular} & Acc.  & \begin{tabular}[c]{@{}c@{}}FLOPs\\ (speedup)\end{tabular} & Acc.  & \begin{tabular}[c]{@{}c@{}}FLOPs\\ (speedup)\end{tabular} \\ \midrule
BERT                                                                               & 95.25 & \begin{tabular}[c]{@{}c@{}}21785M\\ (1.00x)\end{tabular}  & 86.88 & \begin{tabular}[c]{@{}c@{}}21785M\\ (1.00x)\end{tabular}  & 96.84 & \begin{tabular}[c]{@{}c@{}}21785M\\ (1.00x)\end{tabular}  & 86.68 & \begin{tabular}[c]{@{}c@{}}21785M\\ (1.00x)\end{tabular}  & 97.69 & \begin{tabular}[c]{@{}c@{}}21785M\\ (1.00x)\end{tabular}  & 96.71 & \begin{tabular}[c]{@{}c@{}}21785M\\ (1.00x)\end{tabular}  \\ \midrule
\begin{tabular}[c]{@{}c@{}}DistilBERT\\  (6 layers)\end{tabular}                  & 88.58  &  \begin{tabular}[c]{@{}c@{}}10918M\\ (2.00x)\end{tabular}                                     & 83.31 &  \begin{tabular}[c]{@{}c@{}}10918M\\ (2.00x)\end{tabular}                                        & 95.40 &  \begin{tabular}[c]{@{}c@{}}10918M\\ (2.00x)\end{tabular}                                        & 84.12 &  \begin{tabular}[c]{@{}c@{}}10918M\\ (2.00x)\end{tabular}                                        & 97.69 &  \begin{tabular}[c]{@{}c@{}}10918M\\ (2.00x)\end{tabular}                                        & 95.54 &  \begin{tabular}[c]{@{}c@{}}10918M\\ (2.00x)\end{tabular}                                        \\
\begin{tabular}[c]{@{}c@{}}DistilBERT \\ (3 layers)\end{tabular}                  & 87.33 & \begin{tabular}[c]{@{}c@{}}5428M\\ (4.01x)\end{tabular}                                         & 81.17 & \begin{tabular}[c]{@{}c@{}}5428M\\ (4.01x)\end{tabular}                                         & 94.84 & \begin{tabular}[c]{@{}c@{}}5428M\\ (4.01x)\end{tabular}                                         & 84.07 & \begin{tabular}[c]{@{}c@{}}5428M\\ (4.01x)\end{tabular}                                         & 97.58 & \begin{tabular}[c]{@{}c@{}}5428M\\ (4.01x)\end{tabular}                                         & 95.14 & \begin{tabular}[c]{@{}c@{}}5428M\\ (4.01x)\end{tabular}                                         \\
\begin{tabular}[c]{@{}c@{}}DistilBERT \\ (1 layers)\end{tabular}                  & 81.33 & \begin{tabular}[c]{@{}c@{}}1858M\\ (11.72x)\end{tabular}                                        & 77.40 & \begin{tabular}[c]{@{}c@{}}1858M\\ (11.72x)\end{tabular}                                        & 91.35 & \begin{tabular}[c]{@{}c@{}}1858M\\ (11.72x)\end{tabular}                                        & 71.34 & \begin{tabular}[c]{@{}c@{}}1858M\\ (11.72x)\end{tabular}                                        & 96.90 & \begin{tabular}[c]{@{}c@{}}1858M\\ (11.72x)\end{tabular}                                        & 91.13 & \begin{tabular}[c]{@{}c@{}}1858M\\ (11.72x)\end{tabular}                                        \\ \midrule
\begin{tabular}[c]{@{}c@{}}FastBERT \\ (speed=0.1)\end{tabular}                    & 95.25 & \begin{tabular}[c]{@{}c@{}}10741M\\ (2.02x)\end{tabular}  & 86.88 & \begin{tabular}[c]{@{}c@{}}13613M\\ (1.60x)\end{tabular}  & 96.79 & \begin{tabular}[c]{@{}c@{}}4885M\\ (4.45x)\end{tabular}   & 86.59 & \begin{tabular}[c]{@{}c@{}}12930M\\ (1.68x)\end{tabular}  & 97.71 & \begin{tabular}[c]{@{}c@{}}3691M\\ (5.90x)\end{tabular}   & 96.71 & \begin{tabular}[c]{@{}c@{}}3595M\\ (6.05x)\end{tabular}   \\
\begin{tabular}[c]{@{}c@{}}FastBERT \\ (speed=0.5)\end{tabular}                    & 92.00 & \begin{tabular}[c]{@{}c@{}}3191M\\ (6.82x)\end{tabular}   & 86.64 & \begin{tabular}[c]{@{}c@{}}5170M\\ (4.21x)\end{tabular}   & 96.42 & \begin{tabular}[c]{@{}c@{}}2517M\\ (8.65x)\end{tabular}   & 84.05 & \begin{tabular}[c]{@{}c@{}}6352M\\ (3.42x)\end{tabular}   & 97.72 & \begin{tabular}[c]{@{}c@{}}3341M\\ (6.51x)\end{tabular}   & 95.64 & \begin{tabular}[c]{@{}c@{}}1979M\\ (11.00x)\end{tabular}  \\
\begin{tabular}[c]{@{}c@{}}FastBERT\\  (speed=0.8)\end{tabular}                    & 89.75 & \begin{tabular}[c]{@{}c@{}}2315M\\ (9.40x)\end{tabular}   & 85.14 & \begin{tabular}[c]{@{}c@{}}3012M\\ (7.23x)\end{tabular}   & 95.72 & \begin{tabular}[c]{@{}c@{}}2087M\\ (10.04x)\end{tabular}  & 77.45 & \begin{tabular}[c]{@{}c@{}}3310M\\ (6.57x)\end{tabular}   & 97.69 & \begin{tabular}[c]{@{}c@{}}1982M\\ (10.09x)\end{tabular}  & 94.97 & \begin{tabular}[c]{@{}c@{}}1854M\\ (11.74x)\end{tabular}  \\ \midrule \midrule
\multirow{2}{*}{\textbf{\begin{tabular}[c]{@{}c@{}}Dataset/\\ Model\end{tabular}}} & \multicolumn{2}{c|}{\textbf{Ag.news}}                             & \multicolumn{2}{c|}{\textbf{Amz.F}}                                 & \multicolumn{2}{c|}{\textbf{Dbpedia}}                             & \multicolumn{2}{c|}{\textbf{Yahoo}}                               & \multicolumn{2}{c|}{\textbf{Yelp.F}}                              & \multicolumn{2}{c}{\textbf{Yelp.P}}                              \\
                                                                                   & Acc.  & \begin{tabular}[c]{@{}c@{}}FLOPs\\ (speedup)\end{tabular} & Acc.  & \begin{tabular}[c]{@{}c@{}}FLOPs\\ (speedup)\end{tabular} & Acc.  & \begin{tabular}[c]{@{}c@{}}FLOPs\\ (speedup)\end{tabular} & Acc.  & \begin{tabular}[c]{@{}c@{}}FLOPs\\ (speedup)\end{tabular} & Acc.  & \begin{tabular}[c]{@{}c@{}}FLOPs\\ (speedup)\end{tabular} & Acc.  & \begin{tabular}[c]{@{}c@{}}FLOPs\\ (speedup)\end{tabular} \\ \midrule
BERT                                                                               & 94.47 & \begin{tabular}[c]{@{}c@{}}21785M\\ (1.00x)\end{tabular}  & 65.50 & \begin{tabular}[c]{@{}c@{}}21785M\\ (1.00x)\end{tabular}  & 99.31 & \begin{tabular}[c]{@{}c@{}}21785M\\ (1.00x)\end{tabular}  & 77.36 & \begin{tabular}[c]{@{}c@{}}21785M\\ (1.00x)\end{tabular}  & 65.93 & \begin{tabular}[c]{@{}c@{}}21785M\\ (1.00x)\end{tabular}  & 96.04 & \begin{tabular}[c]{@{}c@{}}21785M\\ (1.00x)\end{tabular}  \\ \midrule
\begin{tabular}[c]{@{}c@{}}DistilBERT \\ (6 layers)\end{tabular}                  & 94.64 & \begin{tabular}[c]{@{}c@{}}10872M\\ (2.00x)\end{tabular}                                        & 64.05 & \begin{tabular}[c]{@{}c@{}}10872M\\ (2.00x)\end{tabular}                                        & 99.10 & \begin{tabular}[c]{@{}c@{}}10872M\\ (2.00x)\end{tabular}                                        & 76.73 & \begin{tabular}[c]{@{}c@{}}10872M\\ (2.00x)\end{tabular}                                        & 64.25 & \begin{tabular}[c]{@{}c@{}}10872M\\ (2.00x)\end{tabular}                                        & 95.31 & \begin{tabular}[c]{@{}c@{}}10872M\\ (2.00x)\end{tabular}                                         \\
\begin{tabular}[c]{@{}c@{}}DistilBERT\\  (3 layers)\end{tabular}                  & 93.98 & \begin{tabular}[c]{@{}c@{}}5436M\\ (4.00x)\end{tabular}                                         & 63.84 & \begin{tabular}[c]{@{}c@{}}5436M\\ (4.00x)\end{tabular}                                         & 99.05 & \begin{tabular}[c]{@{}c@{}}5436M\\ (4.00x)\end{tabular}                                         & 76.56 & \begin{tabular}[c]{@{}c@{}}5436M\\ (4.00x)\end{tabular}                                         & 63.50 & \begin{tabular}[c]{@{}c@{}}5436M\\ (4.00x)\end{tabular}                                         & 93.23 & \begin{tabular}[c]{@{}c@{}}5436M\\ (4.00x)\end{tabular}                                          \\
\begin{tabular}[c]{@{}c@{}}DistilBERT \\ (1 layers)\end{tabular}                  & 92.88 & \begin{tabular}[c]{@{}c@{}}1816M\\ (12.00x)\end{tabular}                                        & 59.48 & \begin{tabular}[c]{@{}c@{}}1816M\\ (12.00x)\end{tabular}                                        & 98.95 & \begin{tabular}[c]{@{}c@{}}1816M\\ (12.00x)\end{tabular}                                        & 74.93 & \begin{tabular}[c]{@{}c@{}}1816M\\ (12.00x)\end{tabular}                                        & 58.59 & \begin{tabular}[c]{@{}c@{}}1816M\\ (12.00x)\end{tabular}                                        & 91.59 & \begin{tabular}[c]{@{}c@{}}1816M\\ (12.00x)\end{tabular}                                         \\ \midrule
\begin{tabular}[c]{@{}c@{}}FastBERT \\ (speed=0.1)\end{tabular}                    & 94.38 & \begin{tabular}[c]{@{}c@{}}6013M\\ (3.62x)\end{tabular}   & 65.50    & \begin{tabular}[c]{@{}c@{}}21005M\\ (1.03x)\end{tabular}                                                          & 99.28 & \begin{tabular}[c]{@{}c@{}}2060M\\ (10.57x)\end{tabular}  & 77.37 & \begin{tabular}[c]{@{}c@{}}16172M\\ (1.30x)\end{tabular}  & 65.93 & \begin{tabular}[c]{@{}c@{}}20659M\\ (1.05x)\end{tabular}  & 95.99 & \begin{tabular}[c]{@{}c@{}}6668M\\ (3.26x)\end{tabular}   \\
\begin{tabular}[c]{@{}c@{}}FastBERT \\ (speed=0.5)\end{tabular}                    & 93.14 & \begin{tabular}[c]{@{}c@{}}2108M\\ (10.33x)\end{tabular}  & 64.64    & \begin{tabular}[c]{@{}c@{}}10047M\\ (2.16x)\end{tabular}                                                          & 99.05 & \begin{tabular}[c]{@{}c@{}}1854M\\ (11.74x)\end{tabular}  & 76.57 & \begin{tabular}[c]{@{}c@{}}4852M\\ (4.48x)\end{tabular}   & 64.73 & \begin{tabular}[c]{@{}c@{}}9827M\\ (2.21x)\end{tabular}   & 95.32 & \begin{tabular}[c]{@{}c@{}}3456M\\ (6.30x)\end{tabular}   \\
\begin{tabular}[c]{@{}c@{}}FastBERT \\ (speed=0.8)\end{tabular}                    & 92.53 & \begin{tabular}[c]{@{}c@{}}1858M\\ (11.72x)\end{tabular}  & 61.70    & \begin{tabular}[c]{@{}c@{}}2356M\\ (9.24x)\end{tabular}                                                           & 99.04 & \begin{tabular}[c]{@{}c@{}}1853M\\ (11.75x)\end{tabular}  & 75.05 & \begin{tabular}[c]{@{}c@{}}1965M\\ (11.08x)\end{tabular}  & 60.66 & \begin{tabular}[c]{@{}c@{}}2602M\\ (8.37x)\end{tabular}   & 94.31 & \begin{tabular}[c]{@{}c@{}}2460M\\ (8.85x)\end{tabular}   \\  \bottomrule
\end{tabular}
\end{table*}

\begin{figure*}[ht]
\centering
\includegraphics[width=1.0\textwidth]{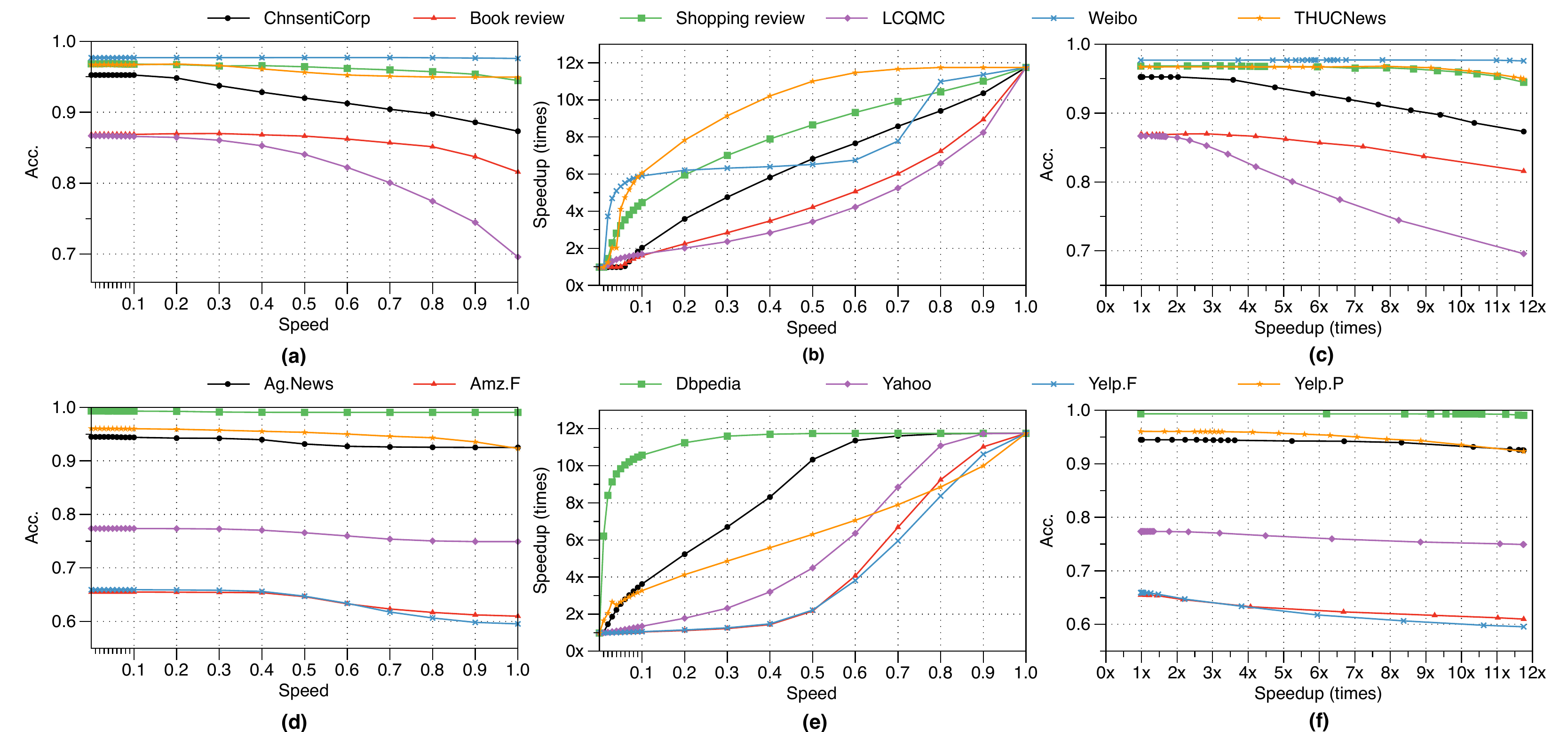}
\caption{The trade-offs of FastBERT on twelve datasets (six in Chinese and six in English): \textbf{(a)} and \textbf{(d)} are \textit{Speed-Accuracy} relations, showing changes of \textit{Speed} (the threshold of \textit{Uncertainty}) in dependence of the accuracy; \textbf{(b)} and \textbf{(e)} are \textit{Speed-Speedup} relations, indicating that the \textit{Speed} manages the adaptibility of FastBERT; \textbf{(c)} and \textbf{(f)} are the \textit{Speedup-Accuracy} relations, i.e. the trade-off between efficiency and accuracy.}
\label{fig:ch_speed_acc}
\end{figure*}

\section{Experimental results}
\label{sec:experimental_results}

In this section, we will verify the effectiveness of FastBERT on twelve NLP datasets (six in English and six in Chinese) with detailed explanations. 

\subsection{FLOPs analysis}
\label{sec:flops_analysis}

Floating-point operations (FLOPs) is a measure of the computational complexity of models, which indicates the number of floating-point operations that the model performs for a single process. The FLOPs has nothing to do with the model's operating environment (CPU, GPU or TPU) and only reveals the computational complexity. Generally speaking, the bigger the model's FLOPs is, the longer the inference time will be. With the same accuracy, models with low FLOPs are more efficient and more suitable for industrial uses.

We list the measured FLOPs of both structures in Table \ref{tab:operation_flops}, from which we can infer that, \textbf{the calculation load (FLOPs) of the Classifier is much lighter than that of the Transformer}. This is the basis of the speed-up of FastBERT, for although it adds additional classifiers, it achieves acceleration by reducing more computation in Transformers.

\subsection{Baseline and dataset}
\label{sec:baseline_dataset}

\subsubsection{Baseline}

In this section, we compare FastBERT against two baselines:

\begin{itemize}
\item \textbf{BERT\footnote{\url{https://github.com/google-research/bert}}} The 12-layer BERT-base model was pre-trained on Wiki corpus and released by Google \citep{devlin2019bert}.
\item \textbf{DistilBERT\footnote{\url{https://github.com/huggingface/transformers/tree/master/examples/distillation}}} The most famous distillation method of BERT with 6 layers was released by Huggingface \citep{sanh2019distilbert}.  In addition, we use the same method to distill the DistilBERT with 3 and 1 layer(s), respectively. 
\end{itemize}

\subsubsection{Dataset}

To verify the effectiveness of FastBERT, especially in industrial scenarios, six Chinese and six English datasets pressing closer to actual applications are used. 
The six Chinese datasets include the sentence classification tasks (ChnSentiCorp, Book review\citep{qiu2018revisiting}, Shopping review, Weibo and THUCNews) and a sentences-matching task (LCQMC\citep{liu2018lcqmc}). All the Chinese datasets are available at the FastBERT project. The six English datasets (Ag.News, Amz.F, DBpedia, Yahoo, Yelp.F, and Yelp.P) are sentence classification tasks and were released in \citep{zhang2015character-level}.

\subsection{Performance comparison}

To perform a fair comparison, BERT / DistilBERT / FastBERT all adopt the same configuration as follows. In this paper, $L=12$. The number of self-attention heads, the hidden dimension of embedding vectors, and the max length of the input sentence are set to 12, 768 and 128 respectively. Both FastBERT and BERT use pre-trained parameters provided by Google, while DistilBERT is pre-trained with \citep{sanh2019distilbert}. We fine-tune these models using the AdamW \citep{ilya2017adamw} algorithm, a $2\times 10^{-5}$ learning rate, and a $0.1$ warmup. Then, we select the model with the best accuracy in 3 epochs. For the self-distillation of FastBERT, we increase the learning rate to $2\times 10^{-4}$ and distill it for 5 epochs.

We evaluate the text inference capabilities of these models on the twelve datasets and report their accuracy (Acc.) and sample-averaged FLOPs under different \textit{Speed} values. The result of comparisons are shown in Table \ref{tab:comparison}, where the $Speedup$ is obtained by using BERT as the benchmark. It can be observed that with the setting of $Speed=0.1$, FastBERT can speed up 2 to 5 times without losing accuracy for most datasets. If a little loss of accuracy is tolerated, FastBERT can be 7 to 11 times faster than BERT. Comparing to DistilBERT, FastBERT trades less accuracy to catch higher efficiency. Figure \ref{fig:ch_speed_acc} illustrates FastBERT's tradeoff in accuracy and efficiency. The speedup ratio of FastBERT are free to be adjusted between 1 and 12, while the loss of accuracy remains small, which is a very attractive feature in the industry.

\begin{figure}[h]
\centering
\includegraphics[width=0.95\columnwidth]{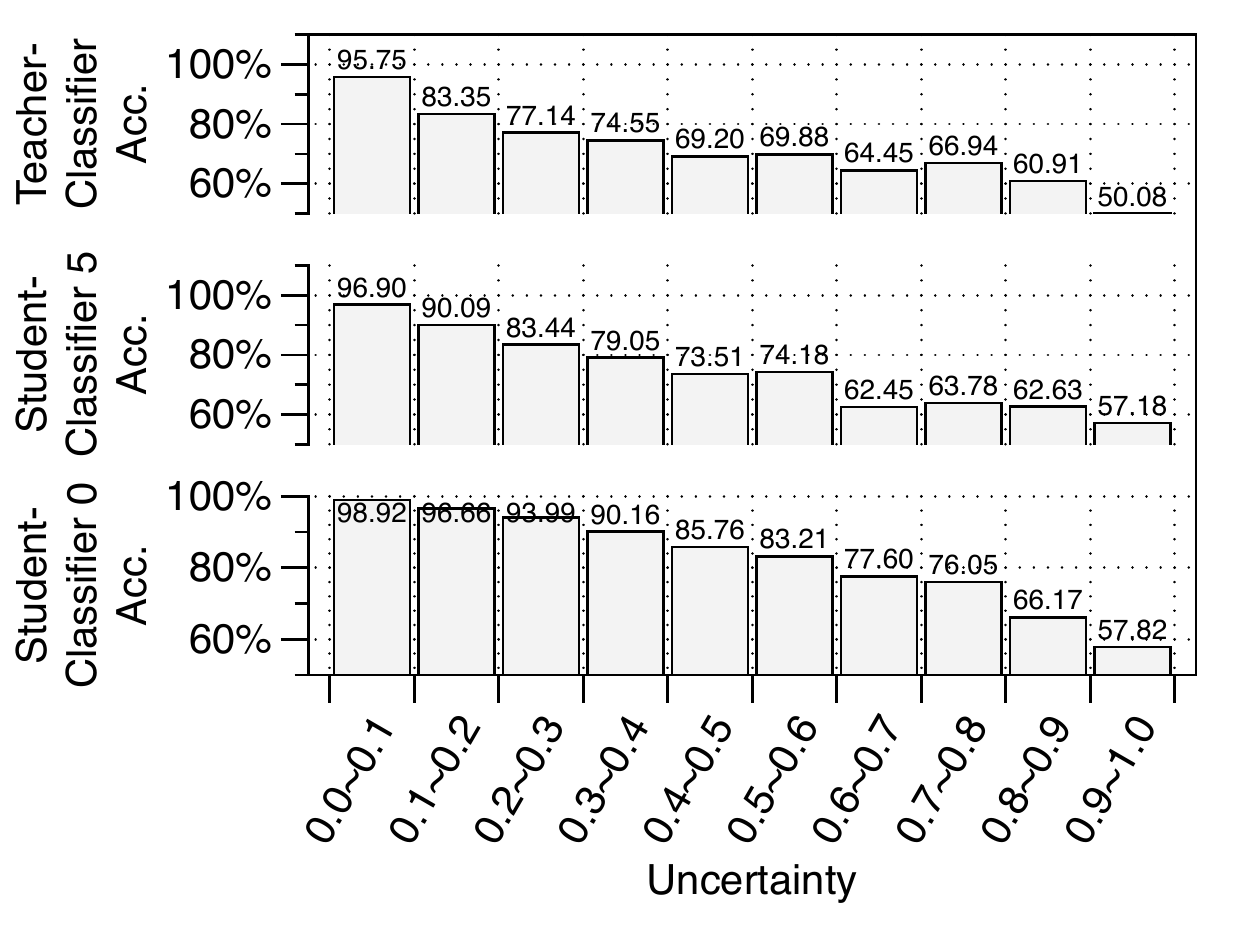}
\caption{The relation of classifier accuracy and average case uncertainty: Three classifiers at the bottom, in the middle, and on top of the FastBERT were analyzed, and their accuracy within various uncertainty intervals were calculated with the Book Review dataset.}
\label{fig:uncertainty_acc}
\end{figure}

\subsection{LUHA hypothesis verification}
\label{hypothesis_verification}

As is described in the Section \ref{sec:self_adaptive_inference}, the adaptive inference of FastBERT is based on the LUHA hypothesis, i.e., ``\textit{the Lower the Uncertainty, the Higher the Accuracy}". Here, we prove this hypothesis using the book review dataset. We intercept the classification results of \textit{Student-Classifier0}, \textit{Student-Classifier5}, and \textit{Teacher-Classifier} in FastBERT, then count their accuracy in each uncertainty interval statistically. As shown in Figure \ref{fig:uncertainty_acc}, the statistical indexes confirm that the classifier follows the LUHA hypothesis, no matter it sits at the bottom, in the middle or on top of the model. 

From Figure \ref{fig:uncertainty_acc}, it is easy to mistakenly conclude that \textit{Student}s has better performance than \textit{Teacher} due to the fact that the accuracy of \textit{Student} in each uncertainty range is higher than that of \textit{Teacher}. This conclusion can be denied by analysis with Figure \ref{fig:uncertainty_distribution}(a) together. For the \textit{Teacher}, more samples are located in areas with lower uncertainty, while the \textit{Student}s' samples are nearly uniformly distributed. Therefore the overall accuracy of the \textit{Teacher} is still higher than that of \textit{Student}s.

\subsection{In-depth study}

In this section, we conduct a set of in-depth analysis of FastBERT from three aspects: the distribution of exit layer, the distribution of sample uncertainty, and the convergence during self-distillation.

\begin{figure}[t]
\centering
\includegraphics[width=0.95\columnwidth]{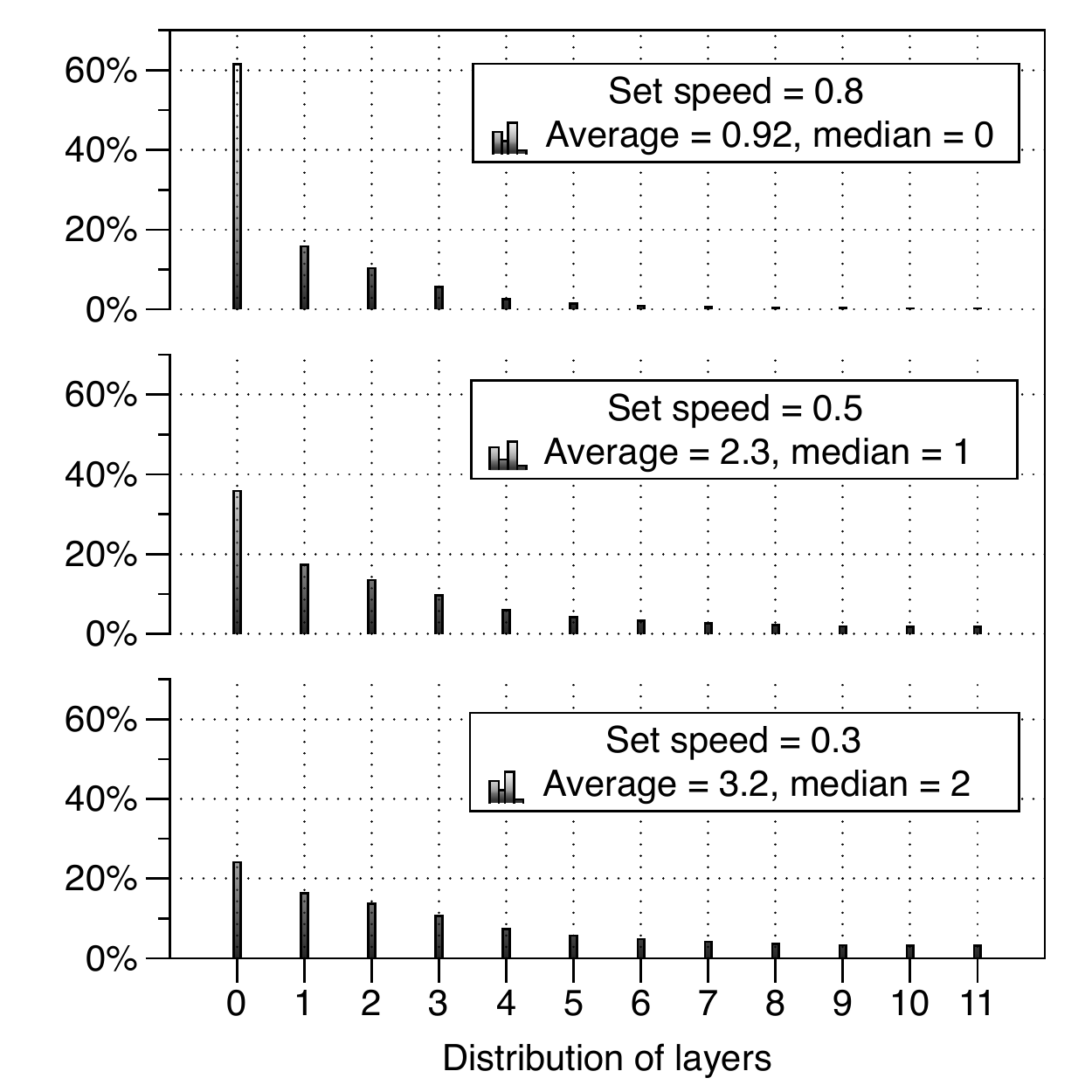}
\caption{The distribution of executed layers on average in the Book review dataset, with experiments at three different speeds (0.3, 0.5 and 0.8).}
\label{fig:layers_distribution}
\end{figure}

\subsubsection{Layer distribution}

In FastBERT, each sample walks through a different number of Transformer layers due to varied complexity. For a certain condition, fewer executed layers often requires less computing resources. As illustrated in Figure \ref{fig:layers_distribution}, we investigate the distribution of exit layers under different constraint of \textit{Speed}s (0.3, 0.5 and 0.8) in the book review dataset. Take $\textit{Speed} = 0.8$ as an example, at the first layer \textit{Transformer0}, 61\% of the samples is able to complete the inference. This significantly eliminates unnecessary calculations in the next eleven layers. 

\subsubsection{Uncertainty distribution}

\begin{figure}[t]
\centering
\includegraphics[width=0.95\columnwidth]{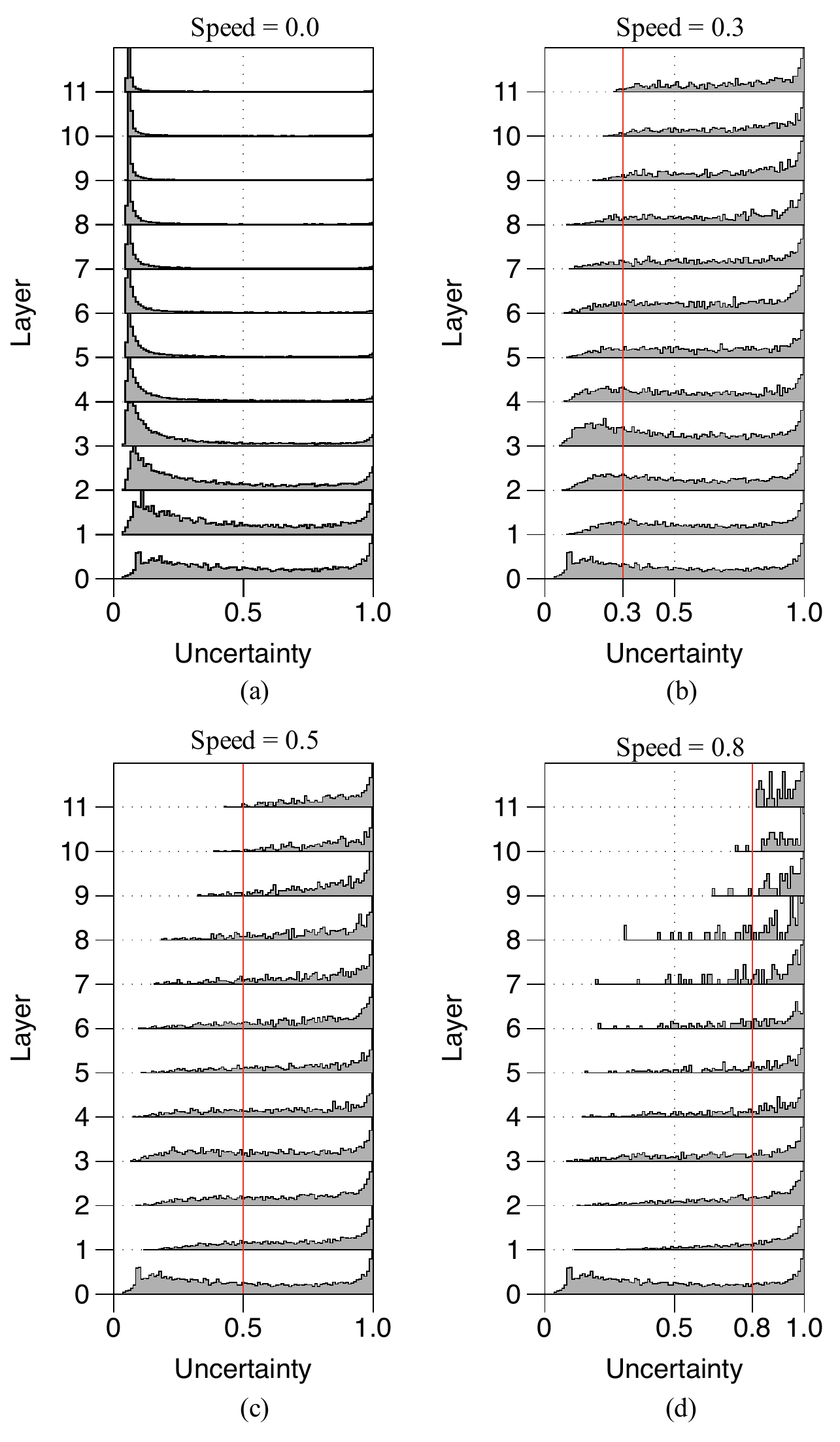}
\caption{The distribution of \textit{Uncertainty} at different layers of FastBERT in the Book review dataset: \textbf{(a)} The \textit{speed} is set to 0.0, which means that all samples will pass through all the twelve layers; \textbf{(b)} $\sim$  \textbf{(d)}: The \textit{Speed} is set to 0.3, 0.5, and 0.8 respectively, iand only the samples with \textit{Uncertainty} higher than \textit{Speed} will be sent to the next layer.}
\label{fig:uncertainty_distribution}
\end{figure}

The distribution of sample uncertainty predicted by different student classifiers varies, as is illustrated in Figure \ref{fig:uncertainty_distribution}. Observing these distributions help us to further understand FastBERT. From Figure \ref{fig:uncertainty_distribution}\textbf{(a)}, it can be concluded that the higher the layer is posited, the lower the uncertainty with given \textit{Speed} will be, indicating that the high-layer classifiers more decisive than the lower ones. It is worth noting that at higher layers, there are samples with uncertainty below the threshold of \textit{Uncertainty} (i.e., the \textit{Speed}), for these high-layer classifiers may reverse the previous judgments made by the low-layer classifiers.

\subsubsection{Convergence of self-distillation}

\begin{figure}[tb]
\centering
\includegraphics[width=0.95\columnwidth]{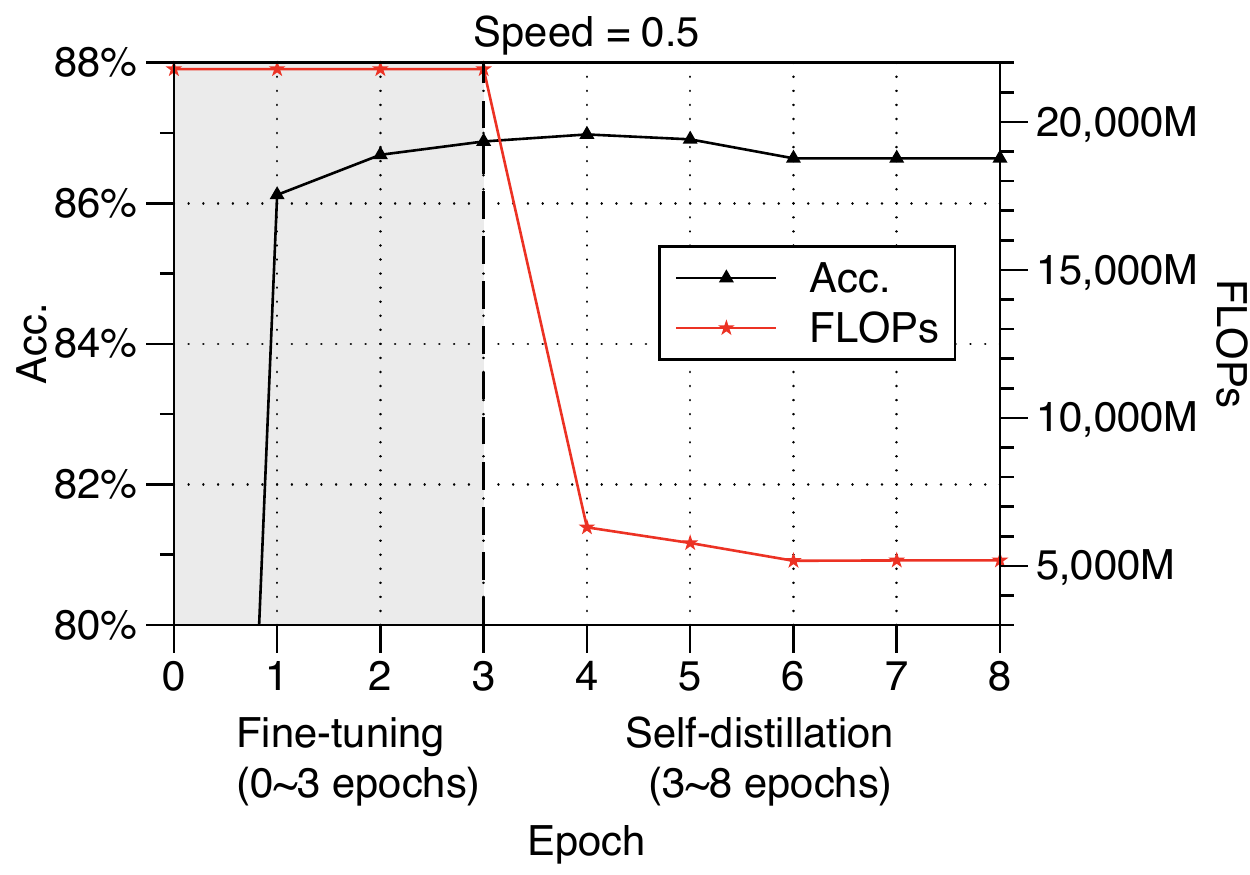}
\caption{The change in accuracy and FLOPs of FastBERT during fine-tuning and self-distillation with the Book review dataset. The accuracy firstly increases at the fine-tuning stage, while the self-distillation reduces the FLOPs by six times with almost no loss in accuracy.}
\label{fig:self_distillation_process}
\end{figure}

Self-distillation is a crucial step to enable FastBERT. This process grants student classifiers with the abilities to infer, thereby offloading work from the teacher classifier. Taking the Book Review dataset as an example, we fine-tune the FastBERT with three epochs then self-distill it for five more epochs. Figure \ref{fig:self_distillation_process} illustrates its convergence in accuracy and FLOPs during fine-tune and self-distillation. It could be observed that the accuracy increases with fine-tuning, while the FLOPs decrease during the self-distillation stage. 

\subsection{Ablation study}

Adaptation and self-distillation are two crucial mechanisms in FastBERT. We have preformed ablation studies to investigate the effects brought by these two mechanisms using the Book Review dataset and the Yelp.P dataset. The results are presented in Table \ref{tab:ablation_study}, in which `without self-distillation' implies that all classifiers, including both the teacher and the students, are trained in the fine-tuning; while `without adaptive inference' means that the number of executed layers of each sample is fixated to two or six.

From Table \ref{tab:ablation_study}, we have observed that: \textbf{(1)} At almost the same level of speedup, FastBERT without self-distillation or adaption performs poorer; \textbf{(2)} When the model is accelerated more than five times, downstream accuracy degrades dramatically without adaption. 
It is safe to conclude that both the adaptation and self-distillation play a key role in FastBERT, which achieves both significant speedups and favorable low losses of accuracy.

\begin{table}[t]
\centering
\small
\caption{Results of ablation studies on the Book review and Yelp.P datasets.}
\label{tab:ablation_study}
\begin{tabular}{c|cc|cc}
\toprule
\multirow{2}{*}{\textbf{Config.}} & \multicolumn{2}{c|}{\textbf{Book review}}                         & \multicolumn{2}{c}{\textbf{Yelp.P}}                              \\
                                  & Acc.  & \begin{tabular}[c]{@{}c@{}}FLOPs\\ (speedup)\end{tabular} & Acc.  & \begin{tabular}[c]{@{}c@{}}FLOPs\\ (speedup)\end{tabular} \\ \midrule
\multicolumn{5}{c}{FastBERT}                                                                                                                                            \\ \midrule
speed=0.2                         & 86.98 & \begin{tabular}[c]{@{}c@{}}9725M\\ (2.23x)\end{tabular}   & 95.90 & \begin{tabular}[c]{@{}c@{}}52783M\\ (4.12x)\end{tabular}  \\
speed=0.7                         & 85.69 & \begin{tabular}[c]{@{}c@{}}3621M\\ (6.01x)\end{tabular}   & 94.67 & \begin{tabular}[c]{@{}c@{}}2757M\\ (7.90x)\end{tabular}   \\ \midrule
\multicolumn{5}{c}{FastBERT without self-distillation}                                                                                                                  \\ \midrule
speed=0.2                         & 86.22 & \begin{tabular}[c]{@{}c@{}}9921M\\ (2.19x)\end{tabular}   & 95.55 & \begin{tabular}[c]{@{}c@{}}4173M\\ (5.22x)\end{tabular}   \\
speed=0.7                         & 85.02 & \begin{tabular}[c]{@{}c@{}}4282M\\ (5.08x)\end{tabular}   & 94.54 & \begin{tabular}[c]{@{}c@{}}2371M\\ (9.18x)\end{tabular}   \\ \midrule
\multicolumn{5}{c}{FastBERT without adaptive inference}                                                                                                                      \\ \hline
layer=6                           & 86.42 & \begin{tabular}[c]{@{}c@{}}11123M\\ (1.95x)\end{tabular}  & 95.18 & \begin{tabular}[c]{@{}c@{}}11123M\\ (1.95x)\end{tabular}  \\
layer=2                           & 82.88 & \begin{tabular}[c]{@{}c@{}}3707M\\ (5.87x)\end{tabular}   & 93.11 & \begin{tabular}[c]{@{}c@{}}3707M\\ (5.87x)\end{tabular}   \\ \bottomrule
\end{tabular}
\end{table}

\section{Conclusion}

In this paper, we propose a fast version of BERT, namely FastBERT. Specifically, FastBERT adopts a self-distillation mechanism during the training phase and an adaptive mechanism in the inference phase, achieving the goal of gaining more efficiency with less accuracy loss. Self-distillation and adaptive inference are first introduced to NLP model in this paper. In addition, FastBERT has a very practical feature in industrial scenarios, i.e., its inference speed is tunable.

Our experiments demonstrate promising results on twelve NLP datasets. Empirical results have shown that FastBERT can be 2 to 3 times faster than BERT without performance degradation. If we slack the tolerated loss in accuracy, the model is free to tune its speedup between 1 and 12 times. Besides, FastBERT remains compatible to the parameter settings of other BERT-like models (e.g., BERT-WWM, ERNIE, and RoBERTa), which means these public available models can be readily loaded for FastBERT initialization.

\section{Future work}

These promising results point to future works in \textbf{(1)} linearizing the \textit{Speed-Speedup} curve; \textbf{(2)} extending this approach to other pre-training architectures such as XLNet \citep{yang2019xlnet} and ELMo \citep{peters2018deep}; \textbf{(3)} applying FastBERT on a wider range of NLP tasks, such as named entity recognition and machine translation.

\section*{Acknowledgments}

This work is funded by 2019 Tencent Rhino-Bird Elite Training Program. Work done while this author was an intern at Tencent.

\bibliography{acl2020}
\bibliographystyle{acl_natbib}

\end{document}